\documentclass{ecai} 
\usepackage{latexsym}
\usepackage{amssymb}
\usepackage{amsmath}
\usepackage{amsthm}
\usepackage{booktabs}
\usepackage{enumitem}
\usepackage{graphicx}
\usepackage{color}

\usepackage{subcaption}
\usepackage[linesnumbered,ruled]{algorithm2e}
\usepackage[capitalise]{cleveref}
\usepackage{mathabx}
\usepackage{complexity}
\usepackage{xspace}
\usepackage{amsfonts}
\usepackage{thm-restate}
\usepackage{tikz}
\usepackage{bm}

\allowdisplaybreaks

\newtheorem{theorem}{Theorem}

\newtheorem{proposition}[theorem]{Proposition}

\newtheorem{definition}{Definition}
\newtheorem{remark}{Remark}

\newcommand{\BibTeX}{B\kern-.05em{\sc i\kern-.025em b}\kern-.08em\TeX}

\newcommand{\sentence}{\Gamma}

\newcommand{\fotwoformula}{\psi}
\newcommand{\formula}{\alpha}

\newcommand{\weight}{w}

\newcommand{\negweight}{\bar{w}}

\newcommand{\world}{\omega}
\newcommand{\wfomc}{WFOMC}

\newcommand{\veck}{\mathbf{k}}

\newcommand{\vecdelta}{\bm{\delta}}

\newcommand{\symwfomc}{\ensuremath{\mathsf{WFOMC}}}

\newcommand{\symwmc}{\ensuremath{\mathsf{WMC}}}

\newcommand{\fotwo}{\ensuremath{\mathbf{FO}^2}\xspace}
\newcommand{\ctwo}{\ensuremath{\mathbf{C}^2}\xspace}

\newcommand{\domain}{\Delta}

\newcommand{\real}{\mathbb{R}}

\newcommand{\fomodels}[2]{\mathcal{M}_{#1, #2}}

\newcommand{\typeweight}[1]{\langle \weight, \negweight\rangle(#1)}

\newcommand{\rlinear}{\widehat{r}}
\newcommand{\wlinear}{\widehat{w}}\newcommand{\rpred}{\widetilde{r}}

\newcommand{\recwfomc}{\textsf{RecursiveWFOMC}}
\newcommand{\incwfomc}{\textsf{IncrementalWFOMC}}
\newcommand{\incwfomctwo}{\textsf{IncrementalWFOMC2}}

\begin{document}

\begin{frontmatter}

\paperid{9592}
\title{Faster Lifting for Ordered Domains with Predecessor Relations}

\author[A]{\fnms{Kuncheng}~\snm{Zou}}
\author[A]{\fnms{Jiahao}~\snm{Mai}}
\author[B]{\fnms{Yonggang}~\snm{Zhang}}
\author[D]{\fnms{Yuyi}~\snm{Wang}} 
\author[E]{\fnms{Ondřej}~\snm{Kuželka}}
\author[A]{\fnms{Yuanhong}~\snm{Wang}\thanks{Corresponding to: lucienwang@jlu.edu.cn, yichang@jlu.edu.cn}}
\author[A,C]{\fnms{Yi}~\snm{Chang}\footnotemark[*]}

\address[A]{School of Artificial Intelligence, Jilin University, China}
\address[B]{College of Computer Science and Technology, Jilin University, China}
\address[C]{Engineering Research Center of Knowledge-Driven Human-Machine Intelligence, MOE, China}
\address[D]{CRRC Zhuzhou Institute, Zhuzhou, China}
\address[E]{Czech Technical University, Czechia}

\begin{abstract}
  We investigate lifted inference on ordered domains with predecessor relations, where the elements of the domain respect a total (cyclic) order, and every element has a distinct (clockwise) predecessor. Previous work has explored this problem through weighted first-order model counting (WFOMC), which computes the weighted sum of models for a given first-order logic sentence over a finite domain. In WFOMC, the order constraint is typically encoded by the linear order axiom introducing a binary predicate in the sentence to impose a linear ordering on the domain elements. The immediate and second predecessor relations are then encoded by the linear order predicate. Although WFOMC with the linear order axiom is theoretically tractable, existing algorithms struggle with practical applications, particularly when the predecessor relations are involved. In this paper, we treat predecessor relations as a native part of the axiom and devise a novel algorithm that inherently supports these relations. The proposed algorithm not only provides an exponential speedup for the immediate and second predecessor relations, which are known to be tractable, but also handles the general $k$-th predecessor relations. The extensive experiments on lifted inference tasks and combinatorics math problems demonstrate the efficiency of our algorithm, achieving speedups of a full order of magnitude. 
\end{abstract}
\end{frontmatter}

\section{Introduction}

Probabilistic inference in relational data, i.e., within the framework of statistical relational models (SRMs)~\citep{VandenBroeck_Kersting_Natarajan_Poole_2021}, is a core task in artificial intelligence.
Among the various inference approaches for SRMs, \emph{lifted} inference with \emph{weighted first-order model counting (WFOMC)}~\citep{vandenbroeckLiftedProbabilisticInference2011} has been widely studied.
In \wfomc{}, an SRM is typically represented as a first-order logic sentence, and the task is to compute the weighted sum of models of the sentence over a finite domain.
For example, an SRM for social networks might encode the smoking habit transition over individuals by $\sentence = \forall x\forall y: sm(x) \land fr(x,y) \Rightarrow sm(y)$, and the \wfomc{} of $\sentence\land sm(x)$ give us the probability of the smoking habit spreading in the network.

In many applications of lifted inference, the domain of interest naturally follows a total order, such as time series data or sequences of events~\citep{le2020probabilistic,vlasselaer2014efficient}.
A common approach in WFOMC to handle such ordered domains is to introduce a \emph{linear order axiom}
% \footnote{\wfomc{} is intracable for first-order logic sentences that contains three logical variables~\citep{Beame_2015}. This prohibits the direct encoding of linear order axioms in first-order logic.} 
in the sentence~\citep{toth2023lifted} that enforces some binary predicate to represent the linear ordering of the domain elements.
For instance, $\forall x\forall y: sm(x) \land older(x,y) \Rightarrow sm(y)$ with a linear order axiom on $older$ means that the smoking habit of an older person can influence a younger person.

\begin{figure}[tbp]
  \centering
  \begin{subfigure}[t]{0.22\textwidth}
    \centering
    \includegraphics[width=\linewidth]{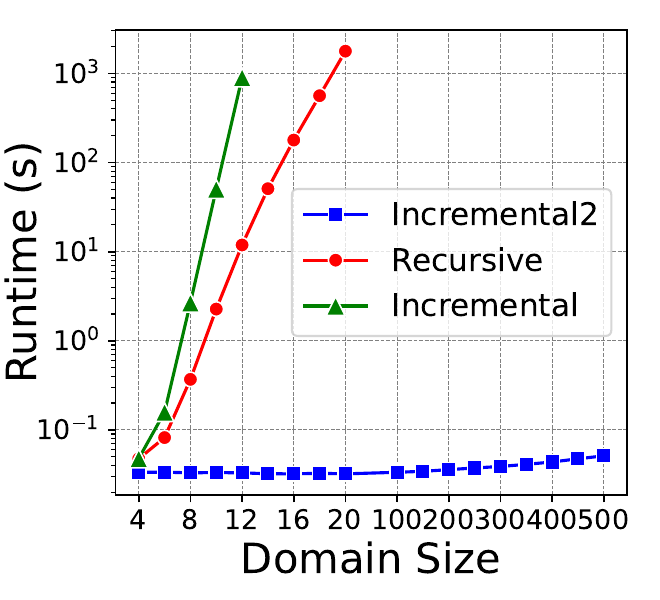}
    \caption{Immediate predecessor}
    \label{fig:immediate_pred}
  \end{subfigure}
  \begin{subfigure}[t]{0.22\textwidth}
    \centering
    \includegraphics[width=\linewidth]{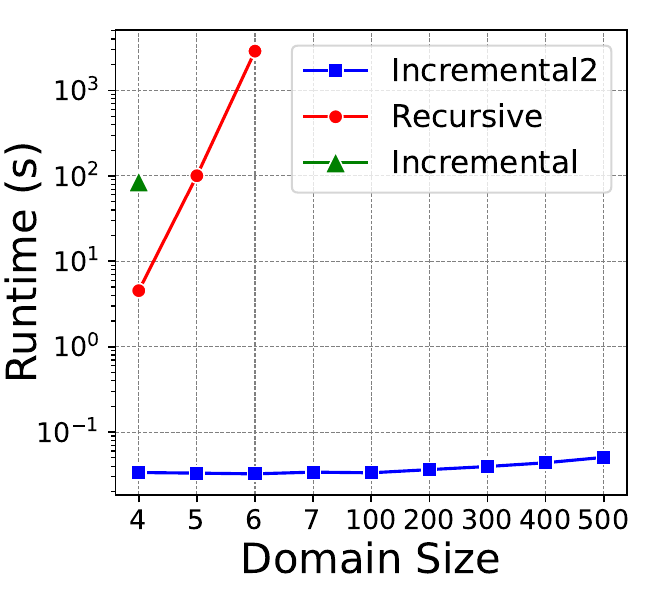}
    \caption{Second predecessor}
    \label{fig:second_pred}
  \end{subfigure}
  \vspace{10pt}
  \caption{Performance comparison of our algorithm (Incremental2) and the previous method (Incremental in \cite{toth2023lifted} and Recursive in \cite{Meng2024}) on the immediate and second predecessor relations.}
  \label{fig:pred_comparison}
\end{figure}

However, the methods to support predecessor relations in \wfomc{} are less explored, though they are essential in most real-world applications.
For example, to represent a hidden Markov model, one needs to encode the transition probabilities between hidden states, which naturally involves the predecessor relations.
Several combinatorial math problems, such as counting permutations with relative position constraints, also require the predecessor relations~\citep{totis2023lifted}. 
\citet{toth2023lifted} proposed to encode the \emph{immediate} and \emph{second} predecessor relations (i.e., predecessor of the immediate predecessor) as derived from the linear order axiom.
Though theoretically tractable, their approach suffers from practical inefficiencies since it relies on complex syntactic extensions of first-order logic language including \emph{counting quantifiers} and \emph{cardinality constraints}.
These extensions are notoriously difficult to handle in practice, usually blowing up the exponent in the polynomial time complexity of the \wfomc{} algorithm~\citep{toth2024complexity}.
For this reason, the authors omitted any experiments on the second predecessor relation in their paper~\cite{toth2023lifted} (from \cref{fig:second_pred}, their algorithm barely scales to the domain size of $4$).
In their subsequent work~\citep{Meng2024}, a more efficient algorithm for the linear order axiom was proposed, while the bottleneck of deriving predecessor relations from the linear order axiom remains.

In this paper, we propose a novel approach to handle the immediate and second predecessor relations in \wfomc{}.
Instead of deriving from the linear order axiom, we treat the predecessor relations as a native part of the axiom, and devise an algorithm that inherently supports these relations.
The algorithm is generalized from the algorithm in \citet{toth2023lifted}, especially exploiting the structure of the predecessor relations with the dynamic programming technique.
Compared to the previous methods, our algorithm is more efficient and straightforward, scaling to hundreds of domain elements for both the immediate and second predecessor relations (see \cref{fig:pred_comparison}).

\begin{figure}[tbp]
  \centering
  \includegraphics[width=.25\textwidth]{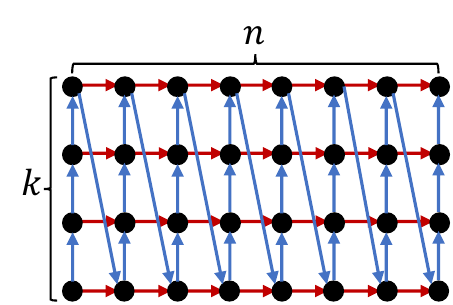}
  \caption{
    The grid is encoded with the general predecessor relation.
  }
  \label{fig:grid}
  % \vspace{2pt}
\end{figure}

Moreover, it turns out surprisingly that our algorithm can be easily extended to handle the (clockwise) \emph{cyclic} predecessor relation and the \emph{general} $k$-th predecessor relation, which is defined as the predecessor of the $(k-1)$-th predecessor.
Particularly, with our algorithm, we prove that the \wfomc{} with the general $k$-th predecessor relation is tractable for the two-variable fragment of first-order logic (\fotwo{}), which was an open problem in the previous work~\citep{toth2023lifted}.
While it is of theoretical interest, we admit that the general $k$-th predecessor relation is not commonly studied in the literature; however, we remark that one of its potential applications to encode the \emph{grid} structure in the \wfomc{} framework would have a significant impact.
With the general predecessor relation, we can encode a $k\times n$ grid with a linear order axiom, whose immediate predecessor relation represents the vertical adjacency and the $k$-th predecessor relation represents the horizontal adjacency (please refer to \cref{fig:grid} and \cref{sec:appendix_grid} for details).
As a result, many probabilistic inference tasks on grids, e.g., the 2-dimensional Ising model with constant interaction strength, which have been shown to be equivalent to the \wfomc{} problem~\citep{VandenBroeck_Kersting_Natarajan_Poole_2021}, can be proven to be tractable with our algorithm if the height $k$ of the grid is bounded by a constant.

We evaluate our algorithm on various lifted inference tasks, including the lifted inference benchmarks from the prior work~\citep{toth2023lifted,Meng2024}, as well as the combinatorics math problems involving permutations.
For lifted inference tasks, our algorithm achieves speedups of a full order of magnitude. For combinatorics problems, our algorithm can solve most of the problems within one second, while previous methods can solve a few problems within the same time limit.

\section{Related Work}

The problem of WFOMC has been extensively investigated for the lifted inference in SRMs. 
Theoretical studies have pinpointed the tractability boundary of computing WFOMCs to lie between the fragments of first-order logic with two (\fotwo{}) and three logical variables ($\mathbf{FO}^3$)~\citep{Beame_2015}.
This implies that directly encoding a linear order in first-order logic might render WFOMC intractable as representing the transitivity of the order requires at least three logical variables.

To overcome this limitation, the linear order axiom was introduced serving as a non-logical assertion that enforces the linear order on the domain elements~\citep{toth2023lifted}.
Such techniques of augmenting \wfomc{}s with additional axioms have also been explored in various contexts~\citep{kuusisto2018weighted,van2023lifted,malhotra2023weighted,kuang_bridging_2024}.
Concurrently, another line of research has focused on identifying more precise tractability boundaries for WFOMC midst \fotwo{} and $\mathbf{FO}^3$~\citep{kuzelka2021weighted,Kazemi_Kimmig_Van_Broeck_Poole_2016,Beame_2015}.
Nevertheless, none of these findings have yielded an approach that directly supports predecessor relations in WFOMC.

In the original paper on the linear order axiom~\citep{toth2023lifted}, the authors demonstrated that the immediate and second predecessor relations can be derived from the linear order axiom.
Their method, however, hinges on complex syntactic manipulations of the input sentence. 
This results in WFOMCs that are tractable in theory but inefficient in practice.
In their subsequent work \citep{Meng2024}, a more efficient algorithm for the linear order axiom was proposed. 
Despite this improvement, the fundamental inefficiency of the derivation-based approach persists.

\section{Preliminaries}

% In this section, we introduce the necessary notations and definitions for our work.
% \subsection{Notations}

We use $[n]$ to denote the set $\{1, 2, \dots, n\}$ for any positive integer $n$.
The bold symbol $\mathbf{n}$ denotes a vector, and $\mathbf{n}_i$ denotes the $i$-th element of $\mathbf{n}$.
The notation $|\mathbf{n}|$ represents the sum of all elements in $\mathbf{n}$: $|\mathbf{n}| = \sum_i \mathbf{n}_i$.
The element-wise subtraction of two vectors $\mathbf{a}$ and $\mathbf{b}$ is defined as: $\mathbf{a} - \mathbf{b} = (a_1 - b_1, a_2 - b_2, \dots)$.
We use $\vecdelta_i$ to denote the $i$-th basis vector, where the $i$-th element is $1$ and the rest are $0$.
Given a non-negative integer vector $\mathbf{n}$ with $|\mathbf{n}| = n$, we use $\binom{n}{\mathbf{n}}$ to denote the multinomial coefficient $\frac{n!}{\prod_{i} \mathbf{n}_i!}$.

\subsection{Fist-Order Logic}

We consider the function-free and finite domain fragment of first-order logic.
A first-order logic formula $\formula$ is defined inductively:
\begin{align*}
    \formula ::= &P(x_1, \dots, x_k) \mid \formula \land \formula \mid \formula \lor \formula \mid \neg \formula \mid \\
    &\quad \formula \Rightarrow \formula \mid \formula                 \Leftrightarrow \formula \mid \exists x: \formula \mid \forall x: \formula
\end{align*}
where $P$ is a predicate symbol from a finite predicate vocabulary, $x_1, \dots, x_k$ are logical variables or constants.
% The equality symbol $=$ is a special binary predicate representing the standard equality relation.
% The symbols $\exists$ and $\forall$ are existential and universal quantifiers, respectively, and the logical connectives $\land$, $\lor$, $\neg$, $\Rightarrow$, and $\Leftrightarrow$ are interpreted as usual.
% Each predicate $P$ is associated with an non-negative integer $k$, called the \emph{arity} of $P$, which specifies the number of arguments that $R$ takes.
% We write $P/k$ to denote a predicate $P$ with arity $k$.
We call the formula in the form $P(x_1, \dots, x_k)$ an \emph{atom}, and an atom or its negation a \emph{literal}.
% A variable $x$ in a formula is called \emph{free} if $x$ is not bound by any quantifier.
A formula is called a \emph{sentence} if all variables in the formula are bound by quantifiers.
A formula is called \emph{ground} if it contains no variables.
% A formula without any variables is called a \emph{ground formula}.
% \emph{Grounding} a formula $\formula$ is the process of replacing all variables in $\formula$ with constants in a finite domain $\domain$, and removing the possible quantifiers.

Given a first-order logic formula $\phi$, we write $\mathcal{P}_{\phi}$ for the set of all predicates that appear in $\phi$.
We adopt the \emph{Herbrand semantics} with a finite domain $\domain$.
A Herbrand base $\mathcal{H}$ is the set of all ground atoms that can be formed from the predicates in $\mathcal{P}_{\phi}$ and the constants in $\domain$.
A \emph{possible world} $\world$ is a subset of $\mathcal{H}$.
The satisfaction relation $\world \models \sentence$ is defined as usual.
A \emph{model} of a sentence $\sentence$ is a possible world that satisfies $\sentence$.
We denote $\fomodels{\sentence}{n}$ as the set of all models of a sentence $\sentence$ over the domain $[n]$.

\subsection{Weighted First-Order Model Counting}

The \textit{first-order model counting problem} (FOMC)
% ~\citep{vandenbroeckLiftedProbabilisticInference2011} 
aims to compute the number of models of a given sentence $\sentence$ over a domain of given size $n$, i.e., $|\fomodels{\sentence}{n}|$.
The \textit{weighted first-order model counting problem} introduces a pair of weighting functions $(\weight, \negweight)$: $\mathcal{P}_\sentence \to \real$, with the weight of a possible world $\world$ defined as
% \begin{equation*}
  $\typeweight{\world} := \prod_{l\in \world}\weight(\mathsf{pred}(l)) \cdot \prod_{l\in \mathcal{H}\setminus \world}\negweight(\mathsf{pred}(l))$
% \end{equation*}
where the function $\mathsf{pred}: \mathcal{H}\to \mathcal{P}_\sentence$ maps each atom to its corresponding predicate.

\begin{definition}[Weighted First-Order Model Counting]
    % Let $(\weight, \negweight)$ be weighting functions on a sentence $\sentence$. 
    The \emph{weighted first-order model counting (\wfomc{})} of a sentence $\sentence$ over a domain of size $n$ under the weighting functions $(\weight, \negweight)$ is given by
    \begin{equation*}
      \symwfomc(\sentence, n, \weight, \negweight) := \sum_{\mu\in\fomodels{\sentence}{n}}\typeweight{\mu}.
    \end{equation*}
\end{definition}

% \lnote{remove this paragraph if we need more space.}
% Note that since the weighting functions are defined on the predicate level, all ground atoms of the same predicate get the same weights. 
% For this reason, the notion of \wfomc{} defined here is also referred to as \textit{symmetric \wfomc{}}~\cite{Beame_2015}.
If the sentence $\sentence$ is ground, the \wfomc{} becomes the well-known \textit{weighted model counting}: $\symwmc(\sentence, \weight, \negweight) = \sum_{\mu\models\sentence}\typeweight{\mu}$~\citep{chavira2008probabilistic}.

This paper, as well as prior research, has mainly focused on the \textit{data complexity} of \wfomc{}---the complexity of computing $\symwfomc(\sentence, n, \weight, \negweight)$ when fixing $\sentence$ and $(\weight, \negweight)$, and treating the domain size $n$ as a \emph{unary} input.
A \wfomc{} algorithm is \emph{domain-lifted} (or \emph{lifted}) if it has polynomial-time data complexity.
A sentence or class of sentences that admits a domain-lifted algorithm is called \textit{domain-liftable} (or \textit{liftable}).
% The liftability property is crucial for the scalability of \wfomc{} in real-world applications, as in practice, the sentence and the weighting functions are usually fixed, and the domain size is the only input that varies~\citep{VandenBroeck_Kersting_Natarajan_Poole_2021}.

\subsection{A Domain-Lifted Algorithm}
\label{sec:domain-liftable}

% Building on~\citep{vandenbroeckLiftedProbabilisticInference2011,van2014skolemization},  it was shown that any sentence with \emph{at most two logical variables (\fotwo{})} is liftable.
There have been several lifted algorithms~\citep{Beame_2015,van2021faster,toth2023lifted} for \wfomc{}.
Here we sketch the one proposed by \citet{toth2023lifted} for \fotwo{}, named \incwfomc{}, as it is the most relevant to our work.

By the normalization in~\citep{gradel1997decision} and the technique of eliminating existential quantifiers in~\citep{van2014skolemization}, any sentence in \fotwo{} can be reduced to the form of $\forall x\forall y: \psi(x,y)$, where $\psi(x, y)$ is a quantifier-free formula.
This reduction respects the \wfomc{} value.
% and thus in the following, we focus on \wfomc{} of sentences in the form $\forall x\forall y: \psi(x,y)$.

We first need the concept of \textit{cells} originally introduced in~\citep{van2021faster}.
A \emph{cell} of a first-order formula $\formula$ is a maximally consistent set of literals formed from atoms in $\formula$ using only a single variable $x$.
For instance, consider the formula $A(x) \land R(x,y) \Rightarrow A(y)$.
There are $2^2 \! = \! 4$ possible cells: 
$\{A(x) \land R(x,x)\}$, $\{A(x) \land \neg R(x,x)\}$, $\{\neg A(x) \land R(x,x)\}$, and $\{\neg A(x) \land \neg R(x,x)\}$. 

% \begin{definition}[Cell]
  % \label{def:cell}
  % A cell of a first-order formula $\formula$ is a maximal consistent set of literals formed from atoms in $\formula$ using only a single variable $x$.
% \end{definition}
% \begin{example}
%   \label{ex:cell}
%   Consider the formula 
%   $A(x) \land R(x,y) \Rightarrow A(y)$.
%   There are $2^2 \! = \! 4$ possible cells: 
%   $\{A(x) \land R(x,x)\}$, $\{A(x) \land \neg R(x,x)\}$, $\{\neg A(x) \land R(x,x)\}$, and $\{\neg A(x) \land \neg R(x,x)\}$. 
% \end{example}

Let us consider the \wfomc{} of $\sentence = \forall x\forall y: \psi(x,y)$ over a domain of size $n$ under $(\weight, \negweight)$.
% Denote by $\tau_1, \tau_2, \dots, \tau_p$ all the cells of $\psi(x,y)$, where 
Denote by $p$ the number of cells of $\psi(x,y)$.
Note that $p$ solely depends on the number of distinct predicates in $\mathcal{P}_\sentence$.

For any $h\in [n]$, consider the possible partitions of $[h]$ into $p$ disjoint sets; each partition represents a series of assignments of subsets of $[h]$ to each cell.
Then for each model $\mu$ of $\sentence$ over the domain $[h]$, there is a unique partition $\mathcal{C} = (C_1, C_2, \dots, C_p)$ consistent with $\mu$ on the unary and reflexive binary atoms.
We call $\mathcal{C}$ the \emph{partition} of $\mu$, and $(|C_1|, \dots, |C_p|)$ the \emph{cell configuration} of $\mu$.
Observe that the model $\mu$ is also a model of the ground formula $\Phi_{\mathcal{C}} = \bigwedge_{i,j\in[p]}\bigwedge_{a\in C_i, b\in C_j} \fotwoformula(a,b)$, and thus we can write $\symwfomc(\sentence, h, \weight, \negweight) = \sum_{\mathcal{C}} \symwmc(\Phi_{\mathcal{C}}, \weight, \negweight)$.
% Then for every model of $\sentence$ over the domain $[h]$, there is a unique model of the ground formula $\Phi_{\mathcal{C}} = \bigwedge_{i,j\in[p]}\bigwedge_{a\in C_i, b\in C_j} \fotwoformula(a,b)$.

Since we know the truth values of the unary and reflexive binary atoms in $\Phi_{\mathcal{C}}$ given by the cell assignments, we can simplify the formula $\fotwoformula(x,y)$ by replacing every unary and reflexive binary atom with true or false as appropriate. 
Write $\fotwoformula_{ij}(x,y)$ for the simplified version of $\fotwoformula(x,y)\land \fotwoformula(y,x)$ when $x$ and $y$ belong to $C_{i}$ and $C_{j}$ respectively, which leads to the following reformulation of $\Phi_{\mathcal{C}}$:
\begin{equation}
  \label{eq:wfomc_lineage}
  \begin{aligned}
    \Phi_{\mathcal{C}} = &\bigwedge_{i,j\in[p]: i<j}\left(\bigwedge_{a\in C_i, b\in C_j} \fotwoformula_{ij}(a,b)\right)\land \\
    & \bigwedge_{i\in[p]} \left(\bigwedge_{a,b\in C_i: a < b} \fotwoformula_{ii}(a,b)\land \bigwedge_{c\in C_i} \fotwoformula_{ii}(c,c)\right).
  \end{aligned}
\end{equation}

Now, let us consider the relationship of the above formula for any partition $\mathcal{C}_1 = (C_1, C_2, \dots, C_p)$ and any $\mathcal{C}_2 = (C_1, \dots, C_l \cup \{h+1\}, \dots, C_p)$.
From \cref{eq:wfomc_lineage}, we have that
\begin{equation}
  \begin{aligned}
  \Phi_{\mathcal{C}_2} &= \Phi_{\mathcal{C}_1} \land \bigwedge_{i\in[p]} \bigwedge_{a\in C_i} \fotwoformula_{il}(a,h+1)\land  \fotwoformula_{ll}(h+1,h+1).
  \end{aligned}
  \label{eq:ground_induction}
\end{equation}
Note that each conjunct in the RHS is independent, as they do not share any propositional variables when grounded, and all these conjuncts are also independent of $\Phi_{\mathcal{C}_1}$.
Let us define
\begin{equation}
  \label{eq:r_s_w}
  \begin{aligned}
    w_i &:= \symwmc(\fotwoformula_{ii}(c,c), \weight, \negweight),\\
    r_{ij} &:= \symwmc(\fotwoformula_{ij}(a,b), \weight, \negweight).\\
    % s_i &= \symwmc(\fotwoformula_{i}(a,b), \weight, \negweight)\\
  \end{aligned}
\end{equation}
By the definition of weighting functions in \wfomc{}, we have that the weighted model countings of $\fotwoformula_{ij}(c,c)$ for every $c\in C_i$ are all equal to $w_i$, and the weighted model countings of $\fotwoformula_{ij}(a,b)$ for every $a\in C_i, b\in C_j$ are all equal to $r_{ij}$.
Thus we can compute $\symwmc(\Phi_{\mathcal{C}_2}, \weight, \negweight)$ as 
\begin{equation}
  \begin{aligned}
    \symwmc(&\Phi_{\mathcal{C}_2}, \weight, \negweight) = \symwmc(\Phi_{\mathcal{C}_1}, \weight, \negweight)\cdot w_l \cdot \prod_{i\in[p]} r_{il}^{|C_i|}.
  \end{aligned}
  \label{eq:induction_wmc}
\end{equation}

% Observe that the value of $\symwmc(\Phi_{\mathcal{C}}, \weight, \negweight)$ is essentially the sum of weights over all models of $\Phi_{\mathcal{C}}$ whose interpretations of the unary and reflexive binary atoms are consistent with the cell assignments in $\mathcal{C}$.
% For brevity, we refer to the consistent partition $(C_1, \dots, C_p)$ of a model as its \emph{partition}, and call the cardinality vector $(|C_1|, \dots, |C_p|)$ its \emph{cell configuration}.

Let $T_h(\veck)$ be the sum of weights over models of $\sentence$ whose cell configuration is $\veck$.
Then by the definition of weighting functions again, we have that $T_h(\veck) = \binom{h}{\veck}\cdot \symwmc(\Phi_{\mathcal{C}}, \weight, \negweight)$, where $\mathcal{C}$ is any partition whose configuration is $\veck$.
From the induction \cref{eq:induction_wmc}, we can write 
\begin{equation}
  \label{eq:incremental_wfomc}
  \begin{aligned}
    T_{h+1}(\veck)\! =\! \sum_{l\in[p]} T_h(\veck - \vecdelta_l)\! \cdot\! w_l\! \cdot\! \prod_{i\in[p]} r_{il}^{(\veck - \vecdelta_l)_i}.
  \end{aligned}
\end{equation}
Finally, the \wfomc{} of $\sentence$ over the domain $[n]$ can be obtained by summing up $T_n(\veck)$ over all $\veck$ such that $|\veck| = n$.
\incwfomc{} computes $T_h(\veck)$ in a dynamic programming manner, and the complexity can be proven to be polynomial in the domain size $n$.
% We call the vectors $\vecn$ in the above equation \emph{cell configurations}.

% Given a partition $(S_1,\dots, S_p)$, we refer to the vector of cardinalities of all cells, $\veck = (|S_1|, \dots, |S_p|)$, as a \emph{cell configuration}.
% Summing across the different possible cell configurations and multiplying by a multinomial coefficient to account for the different possible selections of domain elements for a given configuration, we can derive the \wfomc{} as follows, whose complexity is polynomial in the domain size:
% \begin{equation}
%   \begin{split}
%   \symwfomc(\sentence, n, \weight, \negweight &) = \\
%   &\sum_{|\mathbf{k}| = n} \binom{n}{\mathbf{k}} \prod_{\substack{i,j \in [p] \\ i < j}} r_{ij}^{\mathbf{k}_i \mathbf{k}_j} \prod_{i\in [p]} s_{i}^{\binom{\mathbf{k}_i}{2}} w_i^{\mathbf{k}_i}.
% \label{eq:wfomc}
% \end{split}
% \end{equation}
% \begin{equation*}
% \begin{aligned}
%   &\symwfomc(\sentence,\! n,\! \weight,\! \negweight ) =\\ 
%   &\sum_{\substack{\veck\in\mathbb{N}^p:\\ |\mathbf{k}| = n}} \! \binom{n}{\mathbf{k}} \! \prod_{\substack{i,j \in [p]: \\ i < j}} \! r_{ij}^{\mathbf{k}_i \mathbf{k}_j} \! \prod_{i\in [p]} \! r_{ii}^{\binom{\mathbf{k}_i}{2}} w_i^{\mathbf{k}_i} \! .
% % \label{eq:wfomc}
% \end{aligned}
% \end{equation*}

Further using the technique in \citep{kuzelka2021weighted}, \incwfomc{} can be extended to handle the \ctwo{} fragment (the \fotwo{} sentence with \emph{counting quantifiers} $\exists_{=k}$, $\exists_{\le k}$ and $\exists_{\ge k}$) together with \emph{cardinality constraints}.
The algorithm proposed in this paper is based on \incwfomc{}, and thus permits the same extension.
Therefore, in the rest of the paper, we focus on the \fotwo{} fragment, while all the results can also be applied to the \ctwo{} fragment with cardinality constraints.

% The liftability of \ctwo{} with cardinality constraints was proved by first reducing its WFOMC to a WFOMC in \fotwo{}, and then solved by the lifted algorithm for \fotwo{}~\citep{kuzelka2021weighted}. 
% The algorithms proposed in this paper can also utilize the same reduction, and thus in the rest of the paper, we focus on the \fotwo{} fragment.

% \begin{theorem}[{\cite[Theorem 4]{kuzelka2021weighted}}]
%   \label{thm:ctwo}
%   The two-variable fragment of first-order logic with counting quantifiers and cardinality constraints is domain-liftable.
% \end{theorem}

\subsection{Linear order axiom}

% \lnote{add a sentence to introduce linear order axiom. I think you can find it in Honza's paper.}
The \textit{linear order axiom} enforces some relation in the language to introduce a linear (total) ordering on the domain elements~\citep{libkin2004elements}.
% \lnote{simplify this section if space is limited.}
% Specifically, this axiom asserts that for any pair of elements in the domain, one is less than or equal to the other, thus introducing a concept of sequence or order (wherein the pair of elements can potentially be identical).
% Formally, a binary predicate $R$ satisfies the linear order axiom if it is \emph{reflexive} ($\forall x: R(x,x)$), \emph{connected} ($\forall x\forall y: R(x,y)\lor R(y,x)$), \emph{anti-symmetric} ($\forall x\forall y: (R(x,y)\land R(y,x))\Rightarrow (x = y)$), and \emph{transitive} ($\forall x\forall y\forall z: (R(x,y) \land R(y,z))\Rightarrow R(x,z)$).
We denote by $Linear(R)$ the linear order axiom for a binary predicate $R$.
For clarity, we employ the traditional symbol $\leq$ for the linear order predicate in the following content, and write $x \leq y$ to denote that $x$ is less than or equal to $y$.

When incorporating the linear order axiom into \wfomc{}, directly encoding the axiom in first-order logic may not be applicable, as we need three logical variables to express the transitivity property $R(x,y) \land R(y,z) \Rightarrow R(x,z)$, which is beyond the fragment of \fotwo{}.
In \citep{toth2023lifted}, a slightly modified version of \incwfomc{} was proposed to address this issue, which we briefly introduce in the following.
% The authors demonstrated that the complexity of \incwfomc{} on \fotwo{} sentences with the linear order axiom is polynomial in the domain size, and thus proved the domain-liftability of \fotwo{} with the linear order axiom.
% We provide a brief overview of \incwfomc{} in the following.

Any \fotwo{} sentence with the linear order axiom can be transformed into $\sentence = \forall x\forall y: \psi(x,y)\land Linear(\le)$ by the same transformation as for the \fotwo{} fragment.
\citet{toth2023lifted} first observed that the \wfomc{} of $\sentence$ over a domain of size $n$ can be computed by a \wfomc{} over an \emph{ordered} domain of size $n$.
\begin{proposition}[Corollary~$1$ in~\cite{toth2023lifted}]
  \label{pro:wfomc_linear}
  % Let $\sentence=\forall x\forall y: \psi(x,y)\land Linear(\le)$.
  % be a universally quantified sentence in \fotwo{} with the linear order axiom.
  The \wfomc{} of $\sentence = 
 \forall x\forall y: \psi(x,y)\land Linear(\le)$ over a domain of size $n$ under weighting functions $(\weight, \negweight)$ can be computed by $n! \cdot \sum_{\mu\in\fomodels{\sentence}{n}^\le}\typeweight{\mu}$,
  % \begin{equation*}
    % \symwfomc(\sentence, n, \weight, \negweight) = n! \cdot \sum_{\mu\in\fomodels{\sentence}{n}^\le}\typeweight{\mu},
  % \end{equation*}
  where $\fomodels{\sentence}{n}^\le$ denotes the set of models of $\sentence$ over the domain $[n]$, where $\le$ is interpreted as $1\le 2\le \dots \le n$.
\end{proposition}

% With slight modifications, \incwfomc{} can compute $\sum_{\mu\in\fomodels{\sentence}{n}^\le}\typeweight{\mu}$.
Let $\psi_{ij}^\le(x,y) := \psi_{ij}(x,y)\land x\le y$ and let $\psi_{ij}^<(x,y) := \psi_{ij}(x,y) \land (y\le x) \land \neg (x\le y)$.
Define
\begin{equation*}
  \begin{aligned}
    \wlinear_i &:= \symwmc(\fotwoformula_{ii}^\le(c,c), \weight, \negweight),\\
    \rlinear_{ij} &:= \symwmc(\fotwoformula_{ij}^<(a,b), \weight, \negweight),
  \end{aligned}
  % \label{eq:r_weight}
\end{equation*}
and substitute $w_i$ and $r_{ij}$ in \cref{eq:incremental_wfomc} with $\wlinear_i$ and $\rlinear_{ij}$.
% Then we can compute $T_h(\veck)$ by the following equation:
% \begin{equation}
%   \begin{aligned}
%     T_{h+1}(\veck) = \sum_{l\in[p]} T_h(\veck - \vecdelta_l) \cdot \wlinear_l \cdot \prod_{i\in[p]} \rlinear_{il}^{(\veck - \vecdelta_l)_i}.
%   \end{aligned}
%   \label{eq:incremental_wfomc_linear}
% \end{equation}
Then it is easy to check that $T_h(\veck)$ is exactly the sum of weights over all models in $\fomodels{\sentence}{h}^\le$ whose cell configuration is $\veck$, i.e., $\sum_{\mu\in\fomodels{\sentence}{n}^\le}\typeweight{\mu} = \sum_{\veck\in\mathbb{N}^p: |\veck| = n} T_n(\veck)$, which immediately implies the domain-liftability of \fotwo{} with the linear order axiom.
% It is easy to check that the complexity of \incwfomc{} is polynomial in the domain size $n$.
% The full algorithm is presented in~\cref{alg:incwfomc}.
% Please refer to the original paper~\citep{toth2023lifted} for the proof of the following theorem.
% \begin{theorem}[{\cite[Theorem 4]{toth2023lifted}}]
%   \label{thm:linear_order}
%   The two-variable fragment of first-order logic with the linear order axiom is domain-liftable.
% \end{theorem}

% \begin{algorithm}[tbp]
%     \caption{\incwfomc{} for \fotwo{} with linear order axiom}
%     \label{alg:incwfomc}
%     \KwIn{Sentence $\sentence = \forall x\forall y: \psi(x,y)\land Linear(\le)$, domain size $n$, weighting functions $(\weight, \negweight)$}
%     \KwOut{$\symwfomc(\sentence, n, \weight, \negweight)$}
%     Compute $\wlinear_i$ and $\rlinear_{ij}$ using~\cref{eq:r_weight}\;
%     $\forall i\in[n], \forall \veck\in\mathbb{N}^p: T_i[\veck] \leftarrow 0$\;
%     \ForEach{cell $C_j$}{
%       $T_1[\bm{\vecdelta}_j] \leftarrow \wlinear_j$\;
%     }
%     \ForEach{$i=2$ \KwTo $n$}{
%       \ForEach{cell $C_j$}{
%         \ForEach{$(\veck_{old}, W_{old}) \in T_{i-1}$}{
%           $\veck_{new} \leftarrow \veck_{old} + \bm{\vecdelta}_j$\;
%           $T_i[\veck_{new}] \leftarrow T_i[\veck_{new}] + W_{old} \cdot \wlinear_j \cdot \prod_{l\in[p]} \rlinear_{jl}^{(\veck_{old})_l}$\;
%         }
%       }
%     }
%     \Return $\sum_{\veck\in\mathbb{N}^p: |\mathbf{k}|=n} T_n[\mathbf{k}]$\;
% \end{algorithm}

% \subsection{Predecessor Relations}
% \lnote{mentioned the predecessor relation and the encoding in first-order logic.}

\subsection{Predecessor Relations}

With the linear order axiom, it is natural to consider the predecessor relations on the domain elements.
% ask questions with the constraints on the relative positions of the domain elements, such as the \emph{predecessor relations}.
% define more complex relations on the domain elements, e.g., one may ask a natural question: \emph{What is the predecessor of a given element in the domain?} or \emph{What is the $k$-th predecessor of a given element?}
% Formally, given a linear order predicate $\leq$, its \emph{immediate predecessor relation} $Pred_1$ is defined as:
% \begin{equation}
%   \begin{aligned}
%     % \forall x\forall y: &(x < y) \Leftrightarrow ((x \le y) \land \neg (y \le x))\land \\
%     \forall x\forall y: &Pred_1(x,y) \Leftrightarrow \big((x \le y)\\
%     &\land (\neg \exists z: (x \le z) \land (z \le y))\big).
%   \end{aligned}
%   \label{eq:predecessor}
% \end{equation}
% The \emph{$k$-th predecessor relation} $Pred_k$ is defined inductively as:
% \begin{equation}
% \begin{aligned}
%      &\forall x\forall y: Pred_k(x,y)\Leftrightarrow\\ &(\exists z: Pred_{k-1}(x,z) \land Pred_1(z,y)).
%      \end{aligned}
% \end{equation}
In the context of \wfomc{}, we can view the predecessor relations as an extension of the linear order axiom, and write it as $Linear(\leq, Pred_1, Pred_2, \dots, Pred_k)$, where $Pred_i$ is the $i$-th predecessor relation.

Due to the same reason as the linear order axiom, directly encoding the predecessor relation in first-order logic leads to a formula that \ \incwfomc{} does not support.
In \citep{toth2023lifted}, another encoding that is liftable was proposed for the immediate predecessor relation $Pred_1$ using the counting quantifiers and cardinality constraints:
\begin{equation}
    \begin{aligned}
        &\forall x: \neg Perm(x,x) \land \\
        &\forall x\exists_{=1} y: Perm(x,y) \land \forall y\exists_{=1} x: Perm(x,y) \land\\
        &\forall x\forall y: Pred_1(x,y)\Rightarrow Perm(x,y)\land \\
        &\forall x\forall y: Pred_1(x,y)\Rightarrow (x \le y)\land\\
        &|Pred_1| = n-1,
    \end{aligned}
    \label{eq:old_pred1}
\end{equation}
where $Perm/2$ represents a circular permutation of the domain elements, and $|Pred_1|=n-1$ is a cardinality constraint ensuring that there are exactly $n-1$ pairs of elements in the predecessor relation.
Based on this encoding, the authors showed that \fotwo{} with the extended linear order axiom $Linear(\leq, Pred_1)$ is also liftable~\cite[Theorem 6]{toth2023lifted}.
% \begin{proposition}[{\cite[Theorem 6]{toth2023lifted}}]
%     \label{thm:pred1}
%     The two-variable fragment of first-order logic with the linear order axiom and the immediate predecessor relation is domain-liftable.
% \end{proposition}
This result was further extended to  $Pred_2$ with more intricate encoding, while the general case of $k$-th predecessor relation was left as an open problem.
The authors also stressed in the paper that due to the counting quantifiers used in the encoding, \incwfomc{} may \emph{not} be scalable in practice, though it has polynomial-time complexity.

\section{A Faster Algorithm for Linear Order Axiom with Immediate Predecessor Relation}

In this section, we propose a novel \wfomc{} algorithm, called \incwfomctwo{}, that natively supports the extended linear order axiom with immediate predecessor relation, and exhibits a much better time complexity than \incwfomc{}.

% \subsection{General Linear Order Axiom with Predecessor Relations}

% Let us first consider the case where we only have a single predecessor relation $Pred_1$.

% Observe that \cref{eq:wfomc_incremental} essentially presents a dynamic programming form in the sense that computing $T_{h+1}[\veck]$ only requires the values of $T_h[\veck']$ of the previous domain size $h$ such that $\veck' = \veck - \bm{\vecdelta}_t$ for some $t$.
Recall that in \incwfomc{}, $T_h[\veck]$ denotes the weighted sum of models with a given cell configuration $\veck$ over an ordered domain $1\le 2\le \dots \le h$.
We can view the update from $T_h[\veck]$ to $T_{h+1}[\veck]$ as appending a new element $h+1$ to the ordered domain such that $i\le h+1$ for all $i\le h$.
If we further require that $T_h[\veck]$ only accounts for the models consistent with the immediate predecessor relation $Pred_1$, then the update rule can still be used for the \wfomc{} with the immediate predecessor relation.
In this case, we also need to record the cell assignment of the previous element $h$ to ensure that the updated weight only considers the models containing $Pred_1(h, h+1)$.

Let $\fomodels{\sentence}{n}^{\le,Pred_1}$ denote the set of models of $\sentence$ over the domain $[n]$ such that the predicates $\le$ and $Pred_1$ are interpreted as a standard order relation and its immediate predecessor relation, respectively.
We use $T_h(\veck, C_t)$ to denote the weighted sum of models in $\fomodels{\sentence}{h}^{\le,Pred_1}$ such that the element $h$ is assigned to the cell $C_t$ and the cell configuration is $\veck$.
The weights $\wlinear_i$, $\rlinear_{ij}$ are redefined as:
\begin{equation*}
  \begin{aligned}
    \wlinear_i&=\symwmc(\fotwoformula_{i}^\le(c,c)\land \neg Pred_1(c,c), \weight, \negweight),\\
    \rlinear_{ij}&=\symwmc(\fotwoformula_{ij}^<(a,b)\land \neg Pred_1(a,b)\land  \neg Pred_1(b,a), \weight, \negweight).
  \end{aligned}
\end{equation*}
Intuitively, $\rlinear_{ij}$ represents the weights concerning two elements that are not adjacent in the linear order.
A new weight $\rpred_{ij}$ is further introduced for two consecutive elements:
\begin{equation*}
  \begin{aligned}
    \rpred_{ij} = \symwmc(&\fotwoformula_{ij}^<(a,b)\land Pred_1(b,a) \land\neg Pred_1(a,b), \weight, \negweight).
  \end{aligned}
  % \label{eq:r_weight_pred1}
\end{equation*}
Then $T_{h+1}(\veck, C_t)$ can be computed as:
\begin{equation}
  \begin{aligned}
  T_{h+1}(\veck, C_t) = \sum_{l\in[p]} \Bigg(&T_h(\veck-\vecdelta_l, C_l) \cdot \wlinear_{l} \cdot \rlinear_{lt}^{(\veck-\vecdelta_l)_t-1}\cdot \rpred_{lt}\cdot\\
  & \prod_{i\in[p]:i\neq t} \rlinear_{li}^{(\veck-\vecdelta_l)_i}\Bigg).
  \end{aligned}
  \label{eq:wfomc_incremental_pred1}
\end{equation}
Finally, the \wfomc{} of the sentence with the immediate predecessor relation $Pred_1$ can be computed as $n!\cdot \sum_{\veck\in\mathbb{N}^p: |\mathbf{k}|=n}\sum_{t\in[p]} T_n[\mathbf{k}, C_t]$.
The full algorithm is deferred to \Cref{sec:general_linear_order}, where we discuss the general case of $k$-th predecessor relation.

\begin{algorithm*}[tbp]
  \caption{\incwfomctwo{}}
  \label{alg:incwfomc2}
  \DontPrintSemicolon
  \KwIn{$\sentence = \forall x\forall y: \psi(x,y)\land Linear(\le, Pred_1,\dots, Pred_k)$, domain size $n$, weighting functions $(\weight, \negweight)$}
  \KwOut{$\symwfomc(\sentence, n, \weight, \negweight)$}
  Compute $\wlinear_{1,i}, \rlinear_{ij}$ and $\rpred_{ij,s}$ for every $i,j\in[p]$ and $s\in[k]$\\
  $T_1(\bm{\vecdelta}_l, l, \dots, l) \leftarrow \wlinear_t$ for each $l\in[p]$ \label{line:base_case} \tcp*{$k$ copies of $C_l$}
  \ForEach{$i=2$ \KwTo $n$}{ \label{line:main_loop}
    $T_i(\veck, t_1,\dots,t_k) \leftarrow 0$ for every $\veck\in\mathbb{N}^p$, $t_1,\dots,t_k\in[p]$\\
    \ForEach{$l\in[p]$}{ \label{line:cell_loop}
      \ForEach{$(\veck_{old}, t_1, \dots t_k, W_{old}) \in T_{i-1}$}{ \label{line:old_loop}
        $\veck_{new} \leftarrow \veck_{old} + \bm{\vecdelta}_l$\\
        $\veck, W \gets \veck_{old}, W_{old}$ \\
        \ForEach{$s\in[\min(i-1, k)]$}{ \label{line:pred_loop}
          $W \leftarrow W \cdot \rpred_{lt_s,s}$ \tcp*{deal with the $s$-th predecessor}
          $\veck \leftarrow \veck - \bm{\vecdelta}_{t_s}$\tcp*{remove one element from the $s$-th cell}
          \lIf(\tcp*[f]{update the cell assignment for $T_i$}){$s = 1$}{
            $t_s' \leftarrow l$ 
            \textbf{else} $t_s' \leftarrow t_{s-1}$ 
          }
        }
        $T_i(\veck_{new}, t_1',\dots,t_k') \leftarrow T_i(\veck_{new}, t_1',\dots,t_k') + W\cdot \wlinear_l \cdot \prod_{j\in[p]} \rlinear_{lj}^{\veck_j}$ \label{line:pred_update}
      }
    }\label{line:cell_end}
  }
  \Return $n!\cdot \sum_{\veck\in\mathbb{N}^p: |\mathbf{k}|=n}\sum_{t_1,\dots,t_k\in[p]} T_n[\mathbf{k}, t_1,\dots,t_k]$
\end{algorithm*}

The correctness can be shown by a similar argument as for \incwfomc{}.
Informally, we can write a similar relationship between any two partitions $\mathcal{C}_1$ and $\mathcal{C}_2$ in the form of~\cref{eq:ground_induction}: The conjuncts appended to $\Phi_{\mathcal{C}_1}$ contain 
\begin{equation*}
  \fotwoformula^\le_{il}(a,h+1)\land \neg Pred_1(a,h+1)\land \neg Pred_1(h+1,a)
\end{equation*}
for every $a\in C_i$ such that $i\neq t$ and for every $a\in C_t$ such that $a \neq h$, as well as
\begin{align*}
  \fotwoformula^<_{tl}(h, h+1)\land Pred_1(h,h+1)\land \neg Pred_1(h+1,h)\land \\
  \fotwoformula^\le_{tt}(h+1,h+1)\land \neg Pred_1(h+1,h+1).
\end{align*}
As a result, the induction~\cref{eq:induction_wmc} is modified accordingly, and the update rule~\cref{eq:wfomc_incremental_pred1} is derived.

\begin{remark}
  The adaptation from $T_h(\veck)$ to $T_h(\veck, C_t)$ only expands the number of records of $T_h$ from $n^p$ to $p\cdot n^p$, where $p$ is the number of cells, and so does the time complexity of \incwfomctwo{}.
  Compared to the original encoding with counting quantifiers and cardinality constraints in \citep{toth2023lifted}, whose complexity is $O(n^{c\cdot p})$ for some positive integer $c$, our approach has a much better complexity.
\end{remark}

\section{Generalization to Cyclic Predecessor Relation}

% The proposed \incwfomctwo{} algorithm can be further extended to handle the cyclic predecessor relation $CyPred$, which acts the same as $Pred_1$ except that the cyclic predecessor of the first element in the ordered domain is the last element.
% We can record the cell assignment of the first element in $T_h$ and update the induction rule accordingly.

% Compared to directly encoding the cyclic predecessor relation with $Pred_1$, \incwfomctwo{} can achieve a better complexity in practice by the same reason as for $Pred_1$ and $Pred_2$.

In some applications, we may need to consider the (closed) cyclic predecessor relation, where the predecessor of the first element is the last element.
It is straightforward to encode the cyclic predecessor relation $CirPred/2$ with the immediate predecessor relation $Pred_1$:
\begin{equation}
  \begin{aligned}
    &\forall x: First(x) \Leftrightarrow (\forall y: x \le y)\land \forall x: Last(x) \Leftrightarrow (\forall y: y \le x)\land \\
    &\forall x\forall y: CirPred(x,y) \Leftrightarrow Pred_1(x,y)\lor (Last(x)\land First(y)).
  \end{aligned}
  \label{eq:cyclic_pred}
\end{equation}
Another encoding based on \cref{eq:old_pred1} is also possible, where the predicate $Perm$ essentially represents the cyclic predecessor relation.
Both encodings can result in polynomial-time \wfomc{} computation in theory, while they may lead to a higher complexity in practice due to either the auxiliary predicates $First$ and $Last$ or the counting quantifiers and cardinality constraints.

% Following the same idea of handling predecessor relations, we can also extend the algorithm to compute the \wfomc{} of the cyclic predecessor relation.
% Now, the term $T_h$ is extended to incorporate the cell of first element $1$ in the ordered domain $1\le 2\le \dots \le h$.

% \subsection{Generalization to Cyclic Predecessor Relation}

Following the same idea of handling the immediate predecessor relation, we can directly handle the cyclic predecessor relation in the \incwfomctwo{} algorithm without introducing the auxiliary predicates, counting quantifiers or cardinality constraints.
It is achieved by modifying the update rule of \cref{eq:wfomc_incremental_pred1} to incorporate the cyclic predecessor relation.
First, we record the cell assignment of the first element in the ordered domain in $T_h$ making it $T_h(\veck, C_{t_0}, C_1)$.
The computation of $T_h(\veck, C_{t_0}, C_1)$ is performed inductively as follows:
\begin{itemize}
  \item For $h=1$, initialize $T_1$ as $T_1(\vecdelta_l, C_l, C_l) = \wlinear_l$ for every $l\in[p]$;
  \item For $h<n$, the update rule for $T_{h+1}(\veck, C_{t_0}, C_t)$ is the same as~\cref{eq:wfomc_incremental_pred1};
  \item For $h=n$, if $t_0=t$, then
  \begin{align*}
    &T_n(\veck, C_{t_0}, C_t) = \sum_{l\in[p]} \Bigg(T_{n-1}(\veck-\vecdelta_l, C_{t_0}, C_l) \cdot \\
    &\qquad \wlinear_{l} \cdot \rlinear_{lt}^{(\veck-\vecdelta_l)_t-2}\cdot \rpred_{lt}\cdot \rpred_{t_0l}\cdot \prod_{i\in[p]:i\neq t} \rlinear_{li}^{(\veck-\vecdelta_l)_i}\Bigg),
  \end{align*}
  otherwise,
  \begin{align*}
    &T_n(\veck, C_{t_0}, C_t) = \sum_{l\in[p]} \Bigg(T_{n-1}(\veck-\vecdelta_l, C_{t_0}, C_l) \cdot \\
    &\wlinear_{l} \cdot \rlinear_{lt}^{(\veck-\vecdelta_l)_t-1}\cdot \rpred_{lt}\cdot \rlinear_{lt_0}^{(\veck-\vecdelta_l)_{t_0}-1}\rpred_{t_0l}\cdot \prod_{i\in[p]:i\neq t} \rlinear_{li}^{(\veck-\vecdelta_l)_i}\Bigg).
  \end{align*}
\end{itemize}
Note that in the last case, the order of $t_0$ and $l$ in the subscripts of $\rpred$ is reversed for $t_0 = t$ and $t_0 \neq t$, respectively, accounting for the predecessor relation of the first and last elements.
Finally, the \wfomc{} of the sentence with the cyclic predecessor relation $CirPred$ can be computed as $n!\cdot \sum_{\veck\in\mathbb{N}^p: |\mathbf{k}|=n}\sum_{t_0,t\in[p]} T_n[\mathbf{k}, t_0, t]$.

Similar to the case of the immediate predecessor relation, introducing the cyclic predecessor relation in \incwfomctwo{} only increases the complexity by a factor of $p$, where $p$ is the number of cells, which is much better than the complexity of the encoding either \cref{eq:cyclic_pred} or \cref{eq:old_pred1}.

\section{Lifted inference for Linear Order Axiom with General Predecessor Relations}
\label{sec:general_linear_order}

In this section, we extend our algorithm to handle the general linear order axiom $Linear(\le, Pred_1, \dots, Pred_k)$ with $k$-th predecessor relations.

% \subsection{General Predecessor Relations}

% Consider the case $Linear(\le, Pred_k)$.
% Similarly, we use $\fomodels{\sentence}{n}^{\le,Pred_k}$ to denote the set of models of $\sentence$ over the domain $[n]$ where the predicates $\le$, $Pred_k$ are interpreted as a standard order relation and the $k$-th predecessor relations, respectively.
% Let $T_h(\veck, C_t)$ be the weighted sum of models in $\fomodels{\sentence}{h}^{\le,Pred_k}$ such that the element $\max(h-k+1, 1)$ is assigned to the cell $C_t$, and the cell configuration is $\veck$.
% Redefined the weights $\wlinear_i$, $\rlinear_{ij}$ and $\rpred_{ij}$ by replacing $Pred_1$ with $Pred_k$. 
% Then the update rule for $T_{h+1}(\veck, C_t)$ keeps the same form as~\cref{eq:wfomc_incremental_pred1} except that when $h \le k$, it recovers to \cref{eq:incremental_wfomc}.

% The general case can be generalized in the same manner.
We directly present the algorithm in~\cref{alg:incwfomc2} without repeating the verbose explanation, which should be clear from the previous sections.
We use $T_h(\veck, t_1, \dots, t_k)$ to denote the weighted sum of models in the ordered domain such that the element $h-s+1$ is assigned to the cell $C_{t_s}$ for every $s\in[\min(h,k)]$, and the cell configuration is $\veck$.
The weights $\wlinear_i$, $\rlinear_{ij}$ are now defined respectively as:
\begin{align*}
  \wlinear_i&=\symwmc(\fotwoformula_{i}^\le(c,c)\land \bigwedge_{s\in[k]} \neg Pred_s(c,c), \weight, \negweight),\\
  \rlinear_{ij}&=\symwmc(\fotwoformula_{ij}^<(a,b)\land \bigwedge_{s\in[k]} \neg Pred_s(a,b)\land \neg Pred_s(b,a), \weight, \negweight).
\end{align*}
For each $s\in[k]$, we define the weight $\rpred_{ij,s}$ as:
\begin{align*}
  \rpred_{ij,s} = \symwmc(&\fotwoformula_{ij}^<(a,b)\land Pred_s(b,a) \land \neg Pred_s(a,b)\land \\
  &\bigwedge_{s'\in[k]: s' \neq s} \neg Pred_{s'}(a,b)\land \neg Pred_{s'}(b,a), \weight, \negweight).
\end{align*}

\incwfomctwo{} performs dynamic programming similar to \incwfomc{}.
For $h=1$, the algorithm initializes $T_1(\bm{\vecdelta}_l, t_l, \dots, t_l)$ for every $l\in[p]$ (\cref{line:base_case}).
For $h\ge 2$, the value of $T_h(\veck, t_1, \dots, t_k)$ is computed by the pseudo-code from \cref{line:cell_loop} to \cref{line:cell_end}.
We note that when updating $T_h$, only the $s$-th predecessors that already exist in $[h]$ are considered (\cref{line:pred_loop}).

It is easy to show that the algorithm has a complexity of $O(p^{k+1}\cdot n^p)$, where $p$ is the number of cells.
As a result, we have the following theorem closing the open problem in~\citep{toth2023lifted}, whose proof is deferred to the \cref{sec:appendix_proof}.

\begin{theorem}\label{thm:general_linear_order}
  The two-variable fragment of first-order logic with the general linear order axiom $Linear(\le, Pred_1, \dots, Pred_k)$ is domain-liftable.
\end{theorem}

\begin{remark}
  When there are inconsecutive predecessor relations in the linear order axiom, e.g., $Linear(\le, Pred_1, Pred_{10})$, the complexity of \incwfomctwo{} is still $O(p^{k+1}\cdot n^p)$, where $k$ is the highest order of the predecessor relations, as we need to maintain the cell assignment of the last $k$ elements in $T_{h-1}$ in order to obtain the correct $t_k$ for $T_h$ (which is $t_{k-1}$ in $T_{h-1}$).
  % Though, the complexity is polynomial in the domain size $n$ since the input sentence including the orders of the predecessor relations is fixed.
\end{remark}

% \begin{remark}
%   Though both \incwfomc{} and \incwfomctwo{} have polynomial-time complexity, for the single predecessors $Pred_1$ and $Pred_2$, the complexities of \incwfomctwo{} are both $O(p\cdot n^p)$ which is much better than the $O(n^{3p})$ and $O(n^{9p})$ complexities of \incwfomc{}.
% \end{remark}

\section{Experiments}

\begin{figure*}[htbp]
  \centering
  \begin{minipage}[t]{\textwidth}
    \centering
      \begin{subfigure}[t]{0.22\textwidth}
        \centering
        \includegraphics[width=\linewidth]{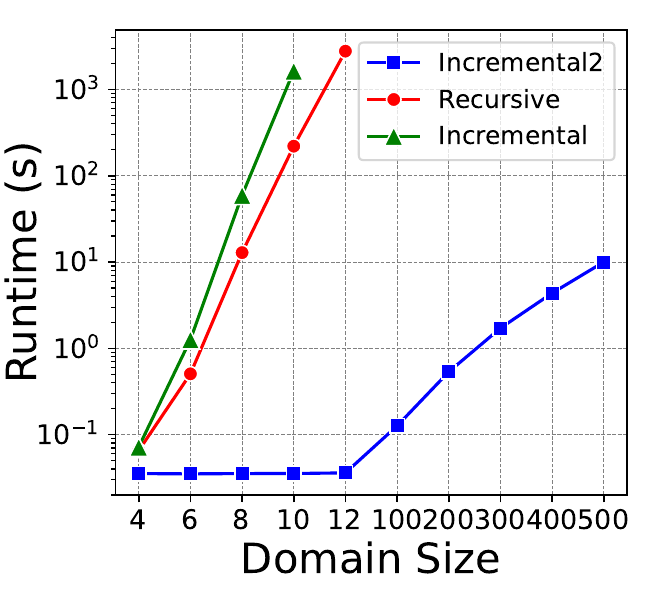}
        \caption{$\sentence_{ws}$ with $m = n$}
        \label{fig:ws1}
      \end{subfigure}
      \begin{subfigure}[t]{0.22\textwidth}
        \centering
        \includegraphics[width=\linewidth]{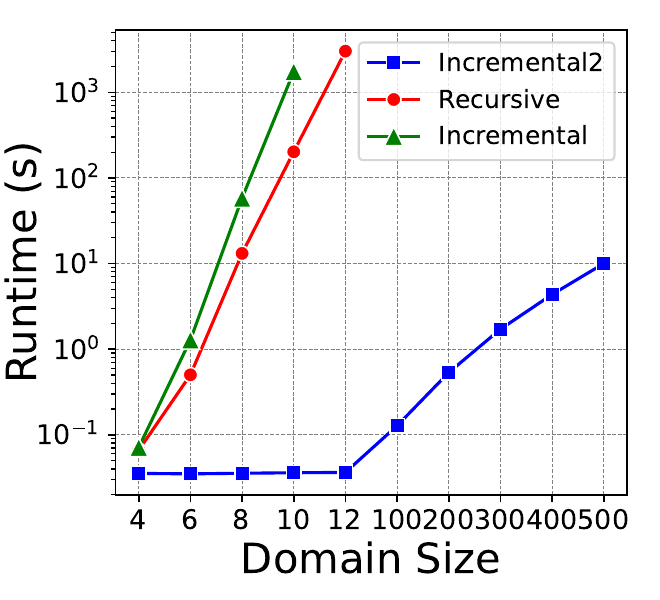}
        \caption{$\sentence_{ws}$ with $m = \left\lceil\frac{3}{4}\right\rceil n$}
        \label{fig:ws2}
      \end{subfigure}
      \begin{subfigure}[t]{0.22\textwidth}
        \centering
        \includegraphics[width=\linewidth]{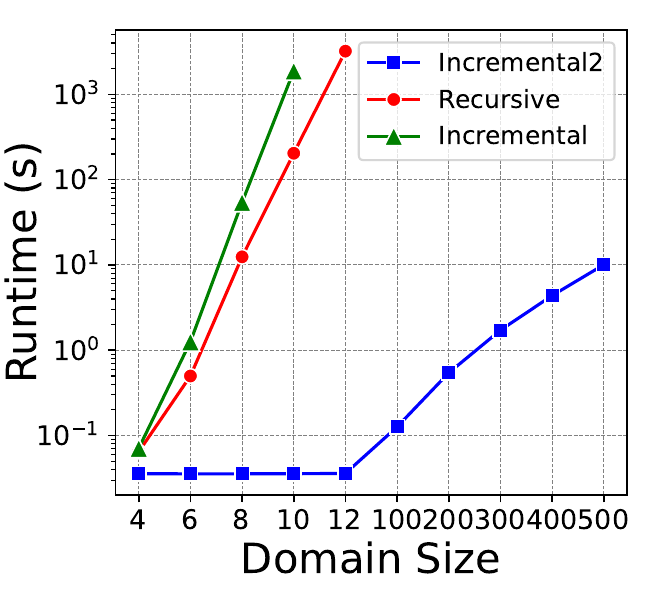}
        \caption{$\sentence_{ws}$ with $m = \frac{1}{2}n$}
        \label{fig:ws3}
      \end{subfigure}
      \begin{subfigure}[t]{0.25\textwidth}
        \includegraphics[width=\linewidth]{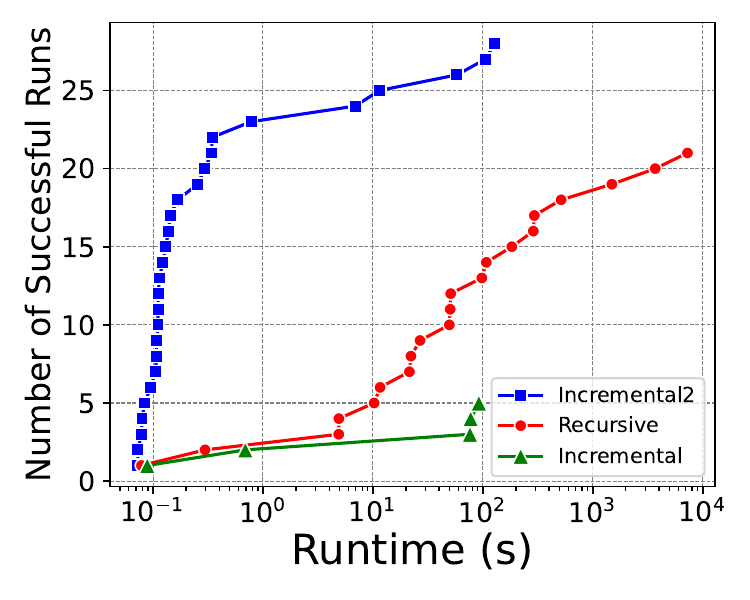}
        \caption{Math problems cactus graph}
        \label{fig:comb_cactus}
      \end{subfigure}
  \end{minipage}
  \newline 
  \vspace{5pt}
  \begin{minipage}[t]{\textwidth}
      \centering
      \begin{subfigure}[t]{0.3\textwidth}
        \centering
        \includegraphics[width=\linewidth]{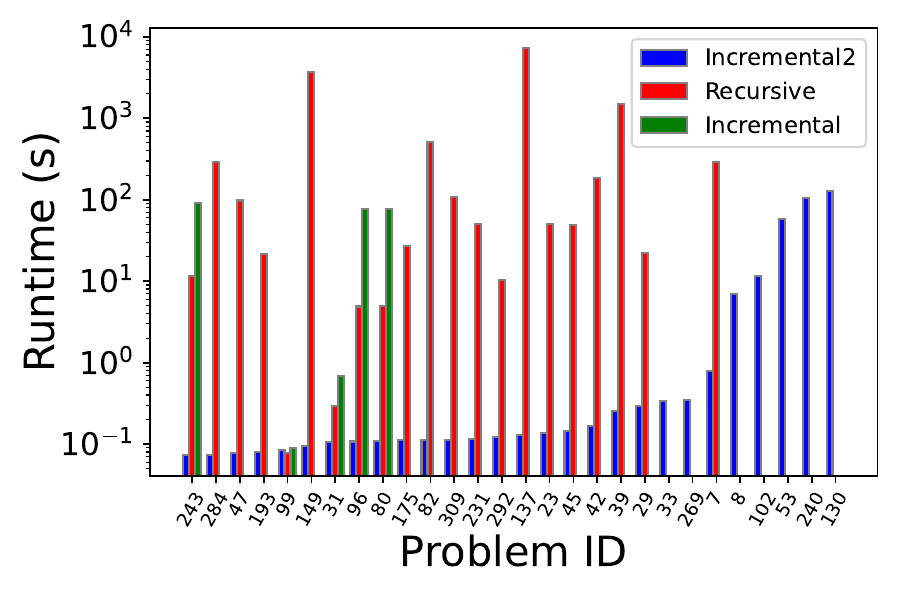}
        \caption{Math problems runtime}
        \label{fig:comb_runtime}
      \end{subfigure}
      \begin{subfigure}[t]{0.22\textwidth}
        \includegraphics[width=\linewidth]{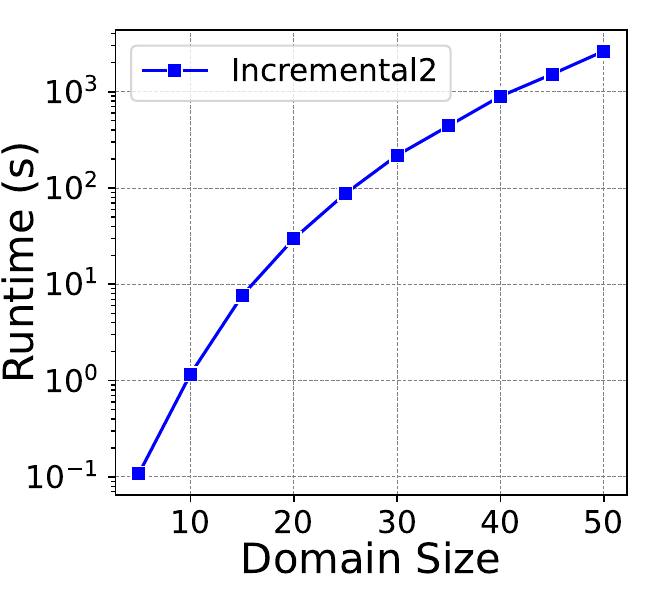}
        \caption{$\sentence_{weather}$}
        \label{fig:mln}
      \end{subfigure}
       \begin{subfigure}[t]{0.22\textwidth}
        \includegraphics[width=\linewidth]{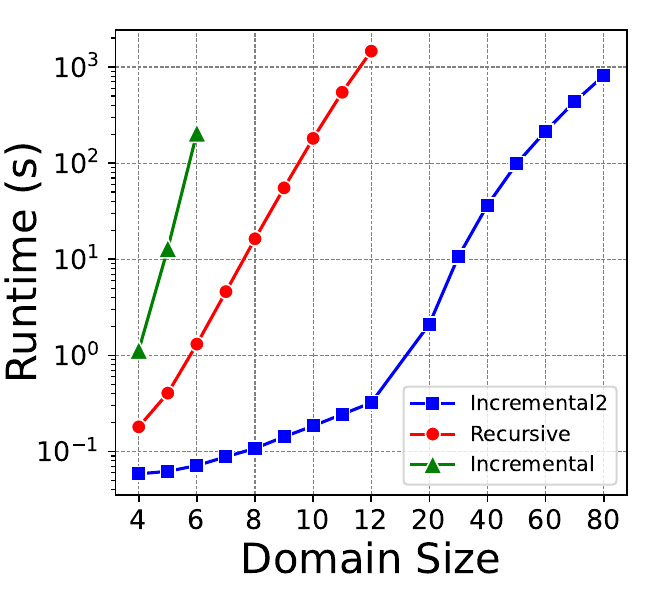}
        \caption{$\sentence_{weather2}$}
        \label{fig:mln2}
      \end{subfigure}
  \end{minipage}
  \vspace{10pt}
  \caption{The results of the experiments}
  \label{fig:experiments}
\end{figure*}

In this section, we compare \incwfomctwo{}
with \incwfomc{}~\citep{toth2023lifted} as well as the state-of-the-art \recwfomc{}~\citep{Meng2024} optimized based on \incwfomc{}.
For \incwfomc{} and \recwfomc{}, we use the encoding-based approach to support the immediate and cyclic predecessor relations as in~\citep{toth2023lifted,Meng2024}.
All experiments were conducted on a machine with four Intel(R) Xeon(R) Gold 5218 CPU and 512GB of RAM.

% We conducted experiments on synthetic problems and combinatorics math problems. 
% Please refer to the appendix for the details of these problems.

\subsection{Watts-Strogatz Model}
\label{sub:ws}

% Due to the native support of \incwfomctwo{} for predecessor relations, we can simplify the problem encoding. We provided both simplified encoding of the problem for IncrementalWFOMC2 and original encoding for RecursiveWFOMC and IncrementalWFOMC, please refer to Appendix for more detail.

% \subsection{Synthetic Problems}
% We synthetically created three types of problems with the linear order axiom and predecessor relations, where the first two are borrowed from~\citep{toth2023lifted}.
% \begin{itemize}
%     \item $\Gamma_{Pred}$: A simple sentence only with the immediate predecessor relation $Pred_1$.
%     The predecessor sentence only encodes the immediate predecessor relation $Pred_1$.
%     The sentence for \incwfomctwo{} is simply
%     \begin{align*}
%     \Gamma_{Pred} = &\forall x: \neg Pred_1(x,x). \\
%     \end{align*}
%     The sentence using the encoding-based approach for \incwfomc{} and \recwfomc{} is
%     \begin{align*}
%     \Gamma_{Pred\_old} = &\forall x: \neg Perm(x,x) \land \\
%     &\forall x \exists y: Perm(x,y) \land \\
%     &\forall y \exists x: Perm(x,y) \land \\
%     &\forall x \forall y: Pred(x,y) \Rightarrow \\ & \ Perm(x,y) \land \\
%     &\forall x \forall y: Pred(x,y) \Rightarrow \\ &\ (x \leq y) \land \\
%     &|Perm| = n \  \land \\ 
%     &|Pred| = n - 1,
%     \end{align*}
%     where $|Perm| = n$ and $|Pred| = n - 1$ are the cardinality constraints.

We first compare the three algorithms on the commonly used synthetic problems, the Watts-Strogatz (WS) model~\citep{watts1998collective}, in the literature~\cite{toth2023lifted,Meng2024}.

The WS model is a procedure for generating random graphs with specific properties.
Given $n$ ordered nodes, we first connect each node to its $K$ closest neighbors (assuming $K$ is an even integer) by undirected edges. 
If the sequence reaches the end or beginning, we wrap around to the other end.
In the experiment, we set $K=2$, resulting in a single cyclic chain that spans all domain elements.
Then, we simply add $m$ additional edges at random to replace the rewired edges in the Watts-Strogatz model.
Consequently, all nodes are connected by the chain, and various shortcuts are also introduced. The sentence $\sentence_{ws}$ is defined as follows
\begin{align*}
    \Gamma_{ws} = &\forall x: \neg  E(x,x) \land \forall x\forall y: E(x,y) \Rightarrow E(y,x)\land \\
    &\forall x\forall y: CirPred(x,y) \Rightarrow E(x,y) \land \\
    &|E| = 2n + 2m.
\end{align*}
We take the same encoding (\cref{eq:old_pred1}) as in~\citep{toth2023lifted,Meng2024} for \incwfomc{} and \recwfomc{} to support the cyclic predecessor relation $CirPred$ (\cref{eq:cyclic_pred}), resulting in the following sentence:
\begin{align*}
    \Gamma_{ws\_old} = &\forall x: \neg Perm(x,x) \land \neg E(x,x) \land \\
    &\forall x \exists y: Perm(x,y) \land  \forall y \exists x: Perm(x,y) \land \\
    &\forall x \forall y: Pred(x,y) \Rightarrow Perm(x,y) \land \\
    &\forall x \forall y: Pred(x,y) \Rightarrow (x \leq y) \land \\
    &\forall x \forall y: Perm(x,y) \Rightarrow E(x,y) \land \\
    &\forall x \forall y: E(x,y) \Rightarrow E(y,x) \land \\
    &|Perm| = n\land |Pred| = n - 1 \land |E| = 2n + 2m.
\end{align*}

The number $m$ of the additional edges is set to $m = n$, $m = \left\lceil\frac{3}{4}\right\rceil n$ and $m = \frac{1}{2}n$ respectively, the same as in~\citep{toth2023lifted,Meng2024}.
The performance of the three algorithms is shown in \cref{fig:ws1,fig:ws2,fig:ws3}.
We can see that \incwfomctwo{} is significantly more efficient than \incwfomc{} and \recwfomc{}.
Particularly, \incwfomctwo{} can complete the computation within a few seconds even for a large domain size (500), while \incwfomc{} and \recwfomc{} are incapable of scaling up to a dozen elements.

\subsection{Combinatorics Math Problems}
\label{sub:math}

To further evaluate the performance of \incwfomctwo{}, we conducted experiments on combinatorics math problems involving the counting of permutations.

We use the MATH dataset~\citep{hendrycksmath2021} to create the benchmark problems.
A total of 28 combinatorics problems that can be encoded with the linear order axiom and the immediate and cyclic predecessor relations were selected from the category of ``counting\_and\_statistics'' in the MATH dataset.

We provide two examples here for illustration.
The first one is: ``How many ways can we put 3 math books and 5 English books on a shelf if all the math books must stay together and all the English books must also stay together?'', and can be encoded with the immediate predecessor relation as:
\begin{align*}
&\forall x: (m(x) \lor e(x)) \land \neg (m(x) \land e(x)) \land \\
&\forall x: mfirst(x)\Leftrightarrow \Big( m(x) \land \neg(\exists y: m(y) \land Pred_1(y,x))\Big) \land \\
&\forall x: efirst(x)\Leftrightarrow \Big( e(x) \land \neg(\exists y: e(y) \land Pred_1(y,x))\Big) \land \\
&|efirst| \le 1 \land |mfirst| \le 1, 
\end{align*}
where $m(x)$ and $e(x)$ denote the book $x$ is a math book and an English book, respectively, and $mfirst$ and $efirst$ denote the first math book and English book on the shelf, respectively.
The cardinality constraints $|efirst| \le 1$ and $|mfirst| \le 1$ ensure that all math books and all English books are together.

Another example of the problem that can be encoded with circular predecessor is: ``In how many ways can we seat 8 people around a table if Alice and Bob won't sit next to each other? ''.
The encoding sentence is:
\begin{align*}
&\forall x\forall y: NextTo(x,y) \Leftrightarrow ( CirPred(x,y) \lor CirPred(y,x) ) \land\\ 
 &\forall x\forall y: (Alice(x) \land Bob(y))\Rightarrow\neg NextTo(x,y)
\end{align*}
where the predicates $Alice$ and $Bob$ interpret the person $x$ as Alice and Bob, respectively.\footnote{When solving this problem with \wfomc{}, we append $Alice(1)$ and $Bob(2)$ to the sentence to explicitly indicate which elements are Alice and Bob.}

All the sentences are transformed for \incwfomc{} and \recwfomc{} with $Pred_1$ and $CirPred$ eliminated by the encoding-based approach.
\cref{fig:comb_cactus} shows the cactus plot of the combinatorics math problems depicting how many problems can be solved within a specified time frame. 
% As shown in the figure, \incwfomctwo{} offers a substantial improvement in runtime compared to \incwfomc{} and \recwfomc{}.
Nearly all the problems can be solved within $100$ seconds by our algorithm, with the majority of them solved within $1$ second.
In contrast, only half and a few problems can be solved by \recwfomc{} and \incwfomc{}, respectively, within the same time frame.
\cref{fig:comb_runtime} further shows the runtimes of all the combinatorics math problems.
From the figure, \recwfomc{} consistently outperforms \incwfomc{} on all problems, but it is still significantly slower than \incwfomctwo{}.

\subsection{Hidden Markov Model with High-Order Dependencies}
\label{sub:weather}

Finally, we evaluate the performance of \incwfomctwo{} for the general $k$-th predecessor relations on a hidden Markov model with high-order dependencies.

The hidden Markov model is represented by a Markov logic network (MLN)~\citep{richardson2006markov} with the following relations:
\begin{align*}
\Psi_{weather} = \{ 
&\infty, \neg (Sn(x) \land Rn(x)), \\
&\infty, Sn(x) \lor Rn(x), \\
&\infty, Sn(x) \land Pred_1(x,y) \Rightarrow Rn(x), \\
&0.5, S(x) \Rightarrow Rn(x), \\
&1.0, S(x) \land Pred_1(x,y)\Rightarrow S(y), \\
&0.4, S(x) \land Pred_2(x,y)\Rightarrow S(y), \\
&0.1, S(x) \land Pred_3(x,y)\Rightarrow S(y). \}
\end{align*}
Intuitively, the weather of someday is either rainy or sunny (formulas 1 and 2); there is a hidden state $S(x)$ indicating the weather of the day $x$ (formula 4); the transition between hidden states is modeled by the last three formulas.
To compute the probability of the rule ``If it is sunny today, it will rain tomorrow'' (formula 3) being true, one might need to compute the partition function of the above MLN.
Using the standard reduction from the partition function of MLNs to the \wfomc{} problem~\cite{vandenbroeckLiftedProbabilisticInference2011}, we have the following \fotwo{} sentence:
 \begin{align*}
    \Gamma_{weather} = 
    &\forall x:\neg (Rn(x) \land Sn(x)) \land(Rn(x) \lor Sn(x)) \land  \\
    &\forall x \forall y: Aux_0(x) \Leftrightarrow (S(x) \Rightarrow Rn(x)) \land \\
    &\forall x \forall y: Pred_1(x,y) \land Sn(x) \Rightarrow Rn(x) \land \\
    &\forall x \forall y: Aux_1(x) \Leftrightarrow (Pred_1(x,y) \land S(x) \Rightarrow S(y) ) \land \\
    &\forall x \forall y: Aux_2(x) \Leftrightarrow (Pred_2(x,y) \land S(x) \Rightarrow S(y) ) \land \\
    &\forall x \forall y: Aux_3(x) \Leftrightarrow (Pred_3(x,y) \land S(x) \Rightarrow S(y) ).
    \end{align*}
The weights are set such that $\weight(Aux_0) = \exp(0.5)$, $\weight(Aux_1) = \exp(1.0)$, $\weight(Aux_2) = \exp(0.4)$, $\weight(Aux_3) = \exp(0.1)$, and $\weight(P) = \negweight(P) = 1$ for all other predicates.
Then the partition function of the above MLN over a domain of size $n$ equals $\symwfomc(\Gamma_{weather}\land Linear(\le, Pred_1, Pred_2, Pred_3), n, \weight, \negweight)$.The runtime of \incwfomctwo{} on $\Gamma_{weather}$ is shown in \cref{fig:mln}, exhibiting a clear sub-exponential growth with the domain size.

Since both \incwfomc{} and \recwfomc{} can neither support the third predecessor relation $Pred_3$ nor scale for the second predecessor relation $Pred_2$, we created another sentence named “weather2” for comparison:
 \begin{align*}
    \Gamma_{weather2} = 
    &\forall x: \neg (Rn(x) \land Sn(x)) \land (Rn(x) \lor Sn(x)) \land  \\
    &\forall x \forall y: Aux_0(x) \Leftrightarrow (S(x) \Rightarrow Rn(x)) \land \\
    &\forall x \forall y: Pred_1(x,y) \land Sn(x) \Rightarrow Rn(x) \land \\
    &\forall x \forall y: Aux_1(x) \Leftrightarrow (Pred_1(x,y) \land S(x) \Rightarrow S(y)).
\end{align*}
The weights are set the same as in $\Gamma_{weather}$ except for $Aux_2$ and $Aux_3$, which are removed from the sentence.

The comparison of the runtime on $\Gamma_{weather2}$ is shown in \cref{fig:mln2}.
As we can see, \incwfomctwo{} showed a significant improvement of a full order of magnitude over \incwfomc{} and \recwfomc{}.

\section{Conclusion}
In this paper, a novel algorithm named IncrementalWFOMC2 is proposed for the \wfomc{} problem with the linear order axiom and predecessor relations.
There are two main contributions: 1) the proposed algorithm natively supports the immediate and cyclic predecessor relations; 2) the algorithm can be extended to support the general linear order axiom with $k$-th predecessor relations, proving that the \fotwo{} fragment with the general linear order axiom is domain-liftable.
Extensive experiments on lifted inference problems and combinatorics math problems demonstrate the significant performance improvement of \incwfomctwo{} over the state-of-the-art algorithms.
In the future, we plan to apply the proposed algorithm to more practical problems, e.g., the sitting arrangement problem in computational economics~\citep{DBLP:conf/wine/BerriaudCW23}.

\begin{ack}
This work was supported by the National Key R\&D Program of China under Grant No.2023YFF0905400 and the National Natural Science Foundation of China (No. U2341229).
Yuyi Wang was supported by the Natural Science Foundation of Hunan Province, China (No.\ 2024JJ5128). 
Ond\v{r}ej Ku{z}elka was supported by the Czech Science Foundation project 23-07299S (``Statistical Relational Learning in Dynamic Domains'').
\end{ack}

% \section*{Acknowledgments}

% \bibliographystyle{named}
\bibliography{mybibfile}

\clearpage

\appendix

\section{Encoding of A Grid with General Predecessor Relations}
\label{sec:appendix_grid}

The basic idea of encoding a $k\times n$ grid with the general predecessor relations is to use the immediate predecessor relations to represent the vertical adjacency relation (i.e., the up relation) $V/2$ and the $k$-th predecessor relations to represent the horizontal adjacency relation (i.e., the right relation) $H/2$.
The corner case where elements are at the boundary of the grid is handled by introducing the auxiliary predicates $Left/1$, $Right/1$, $Top/1$, and $Bottom/1$.

\begin{figure}[htbp]
  \centering
  \includegraphics[width=0.3\textwidth]{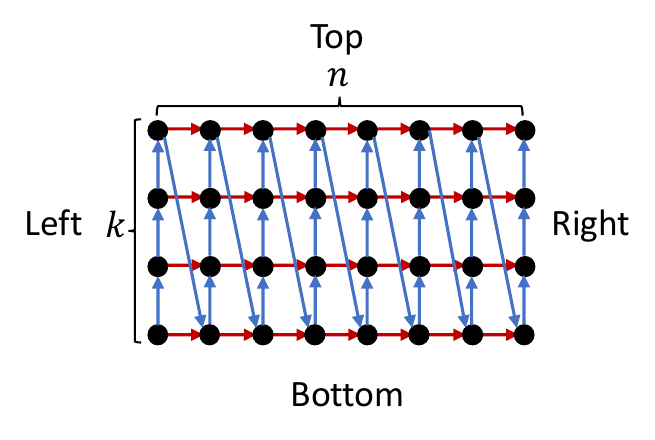}
  \caption{The encoding of a grid with the general predecessor relations.}
  \label{fig:grid_details}
\end{figure}

First, we need the $First/1$ and $Last/1$ predicates to represent the first and last elements in the ordered domain, respectively:
\begin{align*}
  \forall x: &First(x) \Leftrightarrow (\forall y: x \le y)\land \\
  \forall x: &Last(x) \Leftrightarrow (\forall y: y \le x).
\end{align*}
Then, we can define the top and bottom elements in the grid using the $k$-th predecessor relation:
\begin{align*}
  \forall x: &First(x) \Rightarrow Bottom(x)\land \\
  \forall x\forall y: &Bottom(x) \land Pred_k(x,y) \Rightarrow Bottom(y)\land \\
  \forall x: &Last(x) \Rightarrow Top(x)\land \\
  \forall x\forall y: &Top(x) \land Pred_k(y,x) \Rightarrow Top(y).
\end{align*}
The cardinality constraints are used to ensure that there are exactly $n$ elements in the top and bottom rows:
\begin{equation*}
  |Top| = n \land |Bottom| = n.
\end{equation*}
With the top and bottom elements defined, we can easily define the vertical adjacency relation $V/2$ as:
\begin{align*}
  \forall x\forall y: &V(x,y) \Leftrightarrow \left(\neg Top(x)\land \neg Bottom(y) \land Pred_1(x,y)\right).
\end{align*}

Then, the left and right boundaries of the grid can be defined similarly:
\begin{align*}
  \forall x: &First(x) \Rightarrow Left(x),\\
  \forall x\forall y: &Left(x) \land V(x,y) \Rightarrow Left(y),\\
  \forall x: &Last(x) \Rightarrow Right(x),\\
  \forall x\forall y: &Right(x) \land V(y,x) \Rightarrow Right(y)\\
  \land &|Left| = k \land |Right| = k.
\end{align*}
Finally, the horizontal adjacency relation $H/2$ is defined as:
\begin{align*}
  \forall x\forall y: &H(x,y) \Leftrightarrow \left(\neg Left(y)\land \neg Right(x) \land Pred_k(x,y)\right).
\end{align*}

Note that all the formulas above are in the \fotwo{} fragment.
By \Cref{thm:general_linear_order}, the \wfomc{} of the sentence with an additional grid constraint can be computed in time polynomial in the domain size $n$ if the height $k$ of the grid is bounded by a constant.

\section{Proof of \cref{thm:general_linear_order}}
\label{sec:appendix_proof}
Any \fotwo{} sentence with the general linear order axiom can be transformed into $\sentence = \forall x\forall y: \psi(x,y)\land Linear(\le, Pred_1, \dots, Pred_k)$ by the same transformation as for the \fotwo{} fragment.
Consider the \wfomc{} of $\sentence$ over a domain of size $n$ under weighting functions $(\weight, \negweight)$.

Let $\fomodels{\sentence}{n}^{\le,Pred_1,\dots,Pred_k}$ denote the set of models of $\sentence$ over the domain $[n]$ where the predicates $\le$, $Pred_1, \dots, Pred_k$ are interpreted as a standard order relation and the predecessor relations, respectively.
We show that $T_h(\veck, C_{t_1}, \dots, C_{t_k})$ computed in \cref{alg:incwfomc2} is the weighted sum of models in $\fomodels{\sentence}{h}^{\le,Pred_1,\dots,Pred_k}$ such that the cell configuration is $\veck$ and the element $h-s+1$ is assigned to the cell $C_{t_s}$ for every $s\in[\min(h,k)]$.
% It is proved by induction on $h$. 
% For the base case $h=1$, the claim holds as $T_1(\vecdelta_l, C_l, \dots, C_l) = \wlinear_l$ for every $l\in[p]$.

W.l.o.g., we only focus on the case where $h\ge k$.
Consider any fixed $h\in\{k,\dots, n\}$.
For every $s\in[k]$, denote by
$$\Omega_s = \{(1, 1+s), (2, 2+s), \dots, (h - s, h)\}$$
The set of consecutive pairs of elements in the ordered domain $1\le 2\le \dots \le h$ with a distance of $s$.
Let $\Omega = \bigcup_{s\in[k]} \Omega_s$.
Define 
\begin{align*}
  \fotwoformula_{ij}^{\le,\bot}(x,y) = &\fotwoformula_{ij}^\le(x,y)\land \bigwedge_{s\in[k]} \neg Pred_s(x,y),\\
  \fotwoformula_{ij}^{<,\bot}(x,y) = &\fotwoformula_{ij}^<(x,y)\land \bigwedge_{s\in[k]} \Big(\neg Pred_s(x,y)\land \neg Pred_s(y,x)\Big),
\end{align*}
and for every $s\in[k]$, define
\begin{align*}
  &\fotwoformula^{<, s}_{ij}(x,y) = \fotwoformula_{ij}^{<}(x,y)\land Pred_s(y,x)\land \neg Pred_s(x,y)\land\\
  &\qquad \bigwedge_{t\in[k]: t\neq s} \Big(\neg Pred_t(x,y)\land \neg Pred_t(y,x)\Big).
\end{align*}
Consider a cell partition $\mathcal{C} = (C_1, \dots, C_p)$ of $[h]$.
The ground formula $\Phi_{\mathcal{C}}$ (\cref{eq:wfomc_lineage}) now can be written as
\begin{align*}
  \Phi_{\mathcal{C}} = &\bigwedge_{i,j\in[p]: i<j}\Bigg(\bigwedge_{\substack{a\in C_i, b\in C_j:\\ a < b\land (a,b)\notin \Omega}} \fotwoformula_{ij}^{<,\bot}(a,b)\land \bigwedge_{s\in[k]}\bigwedge_{\substack{a\in C_i, b\in C_j:\\ (a,b)\in \Omega_s}} \fotwoformula_{ij}^{<,s}(a,b)\Bigg)\land \\
  &\bigwedge_{i\in[p]} \Bigg(\bigwedge_{\substack{a\in C_i, b\in C_i:\\ a < b\land \{a,b\}\notin \Omega}} \fotwoformula_{ii}^{<,\bot}(a,b)\land \bigwedge_{s\in[k]}\bigwedge_{\substack{a\in C_i, b\in C_i:\\ \{a,b\}\in \Omega_s}} \fotwoformula_{ii}^{<,s}(a,b)\land \\
  &\qquad\qquad \bigwedge_{c\in C_i} \fotwoformula_{ii}^{\le,\bot}(c,c)\Bigg).
\end{align*}
Similar to \cref{eq:ground_induction}, we can write the relationship between two partitions $\mathcal{C}_1 = (C_1, \dots, C_p)$ and $\mathcal{C}_2 = (C_1, \dots, C_l\cup \{h+1\}, \dots, C_p)$ for every $l\in[p]$ as:
\begin{align*}
  &\Phi_{\mathcal{C}_2} = \Phi_{\mathcal{C}_1}\land \fotwoformula_{ll}^{\le,\bot}(h+1,h+1)\land \\
  &\bigwedge_{i\in[p]}\Bigg(\bigwedge_{\substack{a\in C_i: \\(a,h+1)\notin \Omega}} \fotwoformula_{il}^{<,\bot}(a,h+1)\land \bigwedge_{s\in[k]}\bigwedge_{\substack{a\in C_i:\\ (a,h+1)\in \Omega_s}} \fotwoformula_{il}^{<,s}(a,h+1)\Bigg).
\end{align*}
Notice that each conjunct in the above formula is independent, and all these conjuncts are independent of $\Phi_{\mathcal{C}_1}$.
Define
\begin{align*}
  \wlinear_i &= \symwmc(\fotwoformula_{i}^{\le,\bot}(c,c), \weight, \negweight),\\
  \rlinear_{ij} &= \symwmc(\fotwoformula_{ij}^{<,\bot}(a,b), \weight, \negweight),
\end{align*}
and for every $s\in[k]$, and define
\begin{align*}
  \rpred_{ij,s} &= \symwmc(\fotwoformula_{ij}^{<,s}(a,b), \weight, \negweight).
\end{align*}
The weighted model count of $\Phi_{\mathcal{C}_2}$ can be derived from the weighted model count of $\Phi_{\mathcal{C}_1}$:
\begin{align*}
  \symwmc(\Phi_{\mathcal{C}_2}, &\weight, \negweight) = \symwmc(\Phi_{\mathcal{C}_1}, \weight, \negweight) \cdot \wlinear_l \cdot \\
  &\prod_{i\in[p]}\left( \prod_{\substack{a\in C_i\\ (a,h+1)\notin \Omega}} \rlinear_{il} \cdot \prod_{i\in[p]} \prod_{\substack{a\in C_i\\ (a,h+1)\in \Omega_s}} \rpred_{il,s}\right).
\end{align*}
The above equation directly leads to the update of $T_{h+1}$ from $T_h$ in \cref{alg:incwfomc2} (from \cref{line:pred_loop} to \cref{line:pred_update}).
By the meaning of $T_n(\veck, C_{t_1}, \dots, C_{t_k})$, the correctness of the algorithm follows.

Next, let us consider the complexity of the algorithm.
The outer loop at \cref{line:main_loop} runs $O(n)$ times, and the loop for cells at \cref{line:cell_loop} is in $O(p)$.
The inner loop goes through all possible $(\veck_{old}, C_{t_1}, \dots, C_{t_k}, W_{old})$ in $T_{h-1}$, which is $O(p^k\cdot n^{p-1})$.
Thus, the overall complexity is $O(p^{k+1}\cdot n^p)$, which is polynomial in $n$ ($p$ and $k$ are constants when we consider the data complexity).

\end{document}